\documentclass{article}

\PassOptionsToPackage{numbers, compress}{natbib}

\usepackage[final]{neurips_2024}
\usepackage{listings}
\usepackage{caption}
\usepackage{subcaption} 

\usepackage[utf8]{inputenc} % allow utf-8 input
\usepackage[T1]{fontenc}    % use 8-bit T1 fonts
\usepackage[pagebackref=true,breaklinks=true,colorlinks,citecolor=citecolor,linkcolor=black,bookmarks=false]{hyperref}
\usepackage{url}            % simple URL typesetting
\usepackage{booktabs}       % professional-quality tables
\usepackage{amsfonts}       % blackboard math symbols
\usepackage{nicefrac}       % compact symbols for 1/2, etc.
\usepackage{microtype}      % microtypography
\usepackage{xcolor}         % colors
\usepackage{algorithm}
\usepackage[noend]{algpseudocode}
\usepackage{amsmath,amsfonts,bm}

\def\eqref#1{equation~\ref{#1}}
\def\1{\bm{1}}

\DeclareMathAlphabet{\mathsfit}{\encodingdefault}{\sfdefault}{m}{sl}
\SetMathAlphabet{\mathsfit}{bold}{\encodingdefault}{\sfdefault}{bx}{n}

\newcommand{\R}{\mathbb{R}}

\newcommand{\Cov}{\mathrm{Cov}}
\usepackage{graphicx}
\usepackage[font=small]{caption}
\usepackage{wrapfig}
\usepackage{booktabs}
\usepackage{multirow}
\usepackage{arydshln}
\usepackage[inline]{enumitem}
\definecolor{citecolor}{HTML}{0071bc}

\newcommand{\method}{DynaMo}
\newcommand{\tu}[1]{\underline{#1}}
\newcommand{\tb}[1]{\textbf{#1}}

\title{DynaMo: In-Domain Dynamics Pretraining \\ for Visuo-Motor Control}

\author{
  Zichen Jeff Cui\thanks{Corresponding author. Email: \href{mailto:jeff.cui@nyu.edu}{\texttt{jeff.cui@nyu.edu}}}
  \And
  Hengkai Pan
  \And
  Aadhithya Iyer \\ \\
  New York University 
  \And
  Siddhant Haldar
  \And
  Lerrel Pinto
}

\begin{document}

\maketitle

\begin{abstract}

Imitation learning has proven to be a powerful tool for training complex visuo-motor policies. However, current methods often require hundreds to thousands of expert demonstrations to handle high-dimensional visual observations. A key reason for this poor data efficiency is that visual representations are predominantly either pretrained on out-of-domain data or trained directly through a behavior cloning objective.  In this work, we present \method{}, a new in-domain, self-supervised method for learning visual representations. Given a set of expert demonstrations, we jointly learn a latent inverse dynamics model and a forward dynamics model over a sequence of image embeddings, predicting the next frame in latent space, without augmentations, contrastive sampling, or access to ground truth actions. Importantly, \method{} does not require any out-of-domain data such as Internet datasets or cross-embodied datasets. On a suite of six simulated and real environments, we show that representations learned with \method{} significantly improve downstream imitation learning performance over prior self-supervised learning objectives, and pretrained representations. Gains from using \method{} hold across policy classes such as Behavior Transformer, Diffusion Policy, MLP, and nearest neighbors. Finally, we ablate over key components of \method{} and measure its impact on downstream policy performance. Robot videos are best viewed at \url{https://dynamo-ssl.github.io}.

\end{abstract}

\section{Introduction}
\label{sec:introduction}

Learning visuo-motor policies from human demonstrations is an exciting approach for training difficult control tasks in the real world \cite{lee2024behavior, zhao2023learning, chi2023diffusion, cui2022play, shafiullah2022behavior}. However, a key challenge in such a learning paradigm is to efficiently learn a policy with fewer expert demonstrations. To address this, prior works have focused on learning better visual representations, often by pretraining on large Internet-scale video datasets \cite{chen2021learning, ma2022vip, xiao2022masked, nair2022r3m, parisi2022unsurprising, majumdar2024we}. However, as shown in \citet{dasari2023unbiased}, these out-of-domain representations may not transfer to downstream tasks with very different embodiments and viewpoints from the pretraining dataset.

An alternative to using Internet-pretrained models is to train the visual representations `in-domain' on the demonstration data collected to solve the task~\cite{arunachalam2023holo, cui2022play}. However, in-domain datasets are much smaller than Internet-scale data. This has resulted in the use of domain-specific augmentations~\cite{arunachalam2023holo} to induce representational invariances with self-supervision or to collect larger amounts of demonstrations~\cite{zhao2023learning, brohan2022rt}. The reliance of existing methods on large datasets might suggest that in-domain self-supervised pretraining is ineffective for visuo-motor control, and we might be better with simply training end-to-end. In this work, we argue the contrary -- in-domain self-supervision can be effective with a better training objective that extracts more information from small datasets.

Prevalent approaches for using self-supervision in downstream control often make a bag-of-frames assumption, using contrastive methods \cite{chen2021empirical, grill2020bootstrap} or masked autoencoding \cite{majumdar2024we, xiao2022masked} on individual frames for self-supervision. Most of these approaches ignore a rich supervision signal: action-based causality. Future observations are dependent on past observations, and unobserved latent actions. Can we obtain a good visual representation for control by simply learning the dynamics? In fact, this idea is well-established in neuroscience: animals are thought to possess internal models of the motor apparatus and the environment that facilitate motor control and planning \cite{wolpert1995internal, wolpert1998internal, shidara1993inverse, kitazawa1998cerebellar, miall1996forward, jordan1992forward, flanagan1997role, haruno1998multiple}.

In this work, we present \tb{Dyna}mics Pretraining for Visuo-\tb{Mo}tor Control (\tb{DynaMo}), a new self-supervised method for pretraining visual representations for visuomotor control from limited in-domain data. \method{} jointly learns the encoder with inverse and forward dynamics models, without access to ground truth actions \cite{whitney2019dynamics, brandfonbrener2024inverse}. 

\begin{figure}[!t]
  \centering
  \includegraphics[width=\textwidth]{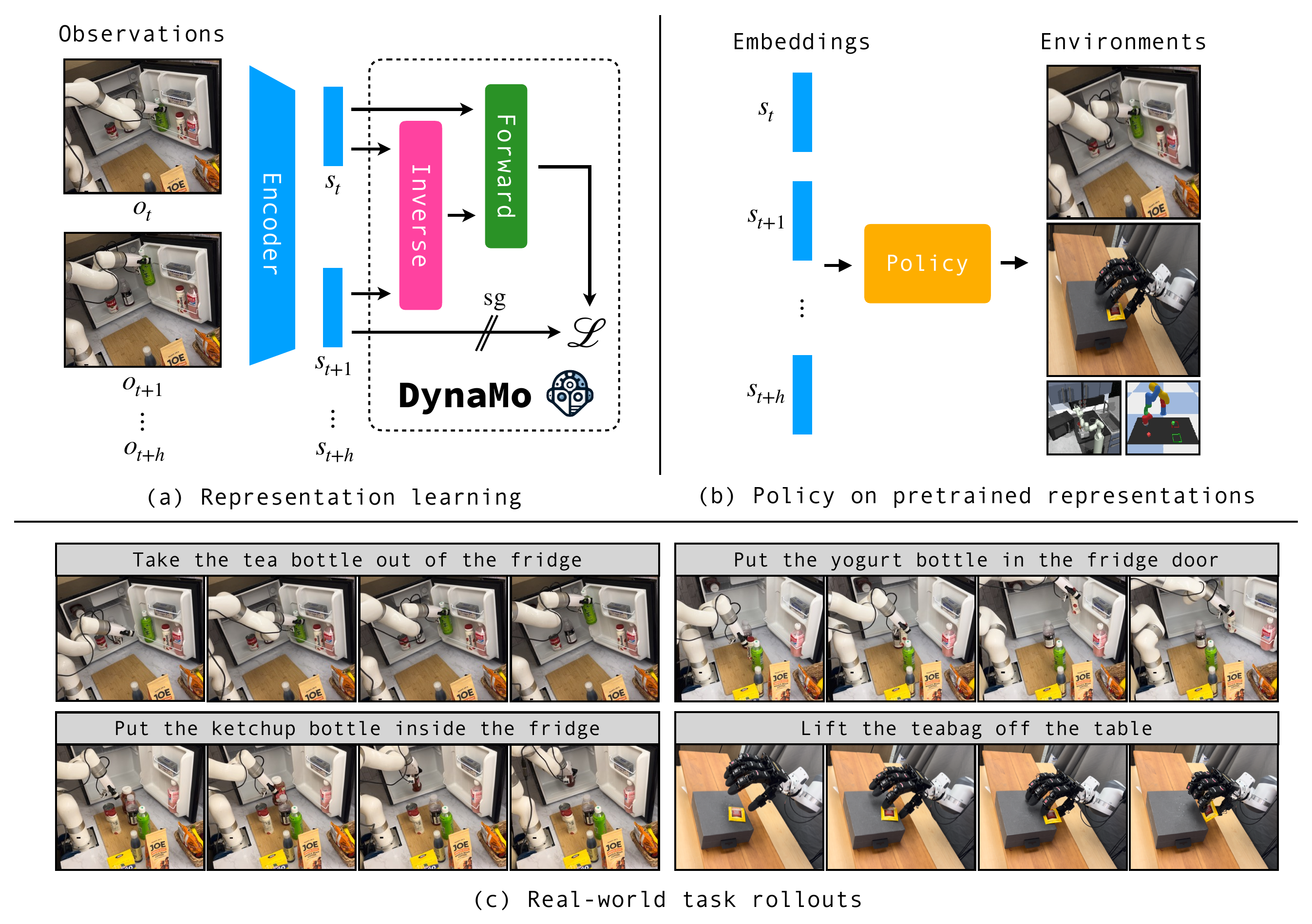}
  \caption{(a) We present \method{}, a new self-supervised method for learning visual representations for visuomotor control. \method{} exploits the causal structure in demonstrations by jointly learning the encoder with inverse and forward dynamics models. \method{} requires no augmentations, contrastive sampling, or access to ground truth actions. This enables downstream policy learning using limited in-domain data across simulated and real-world robotics tasks. For each environment, we pretrain the visual representation in-domain with \method{} and learn a policy on the pretrained embeddings. (b) We provide real-world rollouts of policies learned with \method{} representation on our multi-task xArm Kitchen and Allegro Manipulation environments.}
  \label{fig:intro}
\end{figure}

To demonstrate the effectiveness of \method{}, we evaluate our representation on four simulation suites - Franka Kitchen \cite{gupta2019relay}, Block Pushing \cite{florence2022implicit}, Push-T \cite{chi2023diffusion}, and LIBERO \cite{liu2024libero}, as well as eight robotic manipulation tasks on two real-world environments. Our main findings are summarized below:

\begin{enumerate}[leftmargin=*,align=left]
    \item \method{} exhibits an overall $39\%$ improvement in downstream policy performance over prior state-of-the-art pretrained and self-supervised representations, especially on the harder closed-loop control tasks in Block Pushing and Push-T (Table \ref{tab:sim_rep_results}), and on real robot experiments (Table \ref{tab:real_results}).
    \item \method{} is compatible with various policy classes, can be used to fine-tune pretrained weights, and works in the low-data regime with limited demonstrations on a real-world Allegro hand (Tables \ref{tab:pusht_policies}, \ref{tab:imagenet_finetune}, and \ref{tab:real_results} respectively).
    \item Through an ablation analysis, we study the impact of each component in \method{} on downstream policy performance (\S \ref{exp:ablations}).
\end{enumerate}

All of our datasets, and training and evaluation code will be made publicly available. Videos of our trained policies can be seen here: \url{https://dynamo-ssl.github.io}.
\section{Background}
\label{sec:background}

\subsection{Visual imitation learning}
Our work follows the general framework for visual imitation learning. Given demonstration data $\mathcal{D} = \{(o_t, a_t)\}_t$, where $o_t$ are raw visual observations and $a_t$ are the corresponding ground-truth actions, we first employ a visual encoder $f_\theta: o_t \rightarrow s_t$ to map the raw visual inputs to lower-dimensional embeddings $s_t$. We then learn a policy $\pi(a_t|s_t)$ to predict the appropriate actions. For rollouts, we model the environment as a Markov Decision Process (MDP), where each subsequent observation $o_{t+1}$ depends on the previous observation-action pair $(o_t, a_t)$. We assume the action-conditioned transition distribution $p(o_{t+1}|o_t, a_t)$ to be unimodal for our manipulation tasks.

\subsection{Visual pretraining for policy learning}

Our goal is to pretrain the visual encoder $f_\theta$ using a dataset of sequential raw visual observations $\mathcal{D} = \{o_t\}_t$ to support downstream policy learning. During pretraining, we do not assume access to the ground-truth actions $\{a_t\}_t$.

Prior work has shown that pretraining encoders on large out-of-domain datasets can improve downstream policy performance~\cite{chen2021learning, ma2022vip, xiao2022masked, nair2022r3m, parisi2022unsurprising, majumdar2024we}. However, such pretraining may not transfer well to tasks with different robot embodiments~\cite{dasari2023unbiased}.

Alternatively, we can directly pretrain the encoder in-domain using self-supervised methods. One approach is contrastive learning with data augmentation priors, randomly augmenting an image twice and pushing their embeddings closer. Another approach is denoising methods, predicting the original image from a noise-degraded sample (e.g. by masking \cite{majumdar2024we, xiao2022masked, he2022masked}). A third approach is contrastive learning with temporal proximity as supervision, pushing temporally close frames to have similar embeddings \cite{sermanet2018time, young2022playful}.

\section{\method{}}
\label{sec:approach}

\begin{figure}[h]
  \centering
  \includegraphics[width=\textwidth]{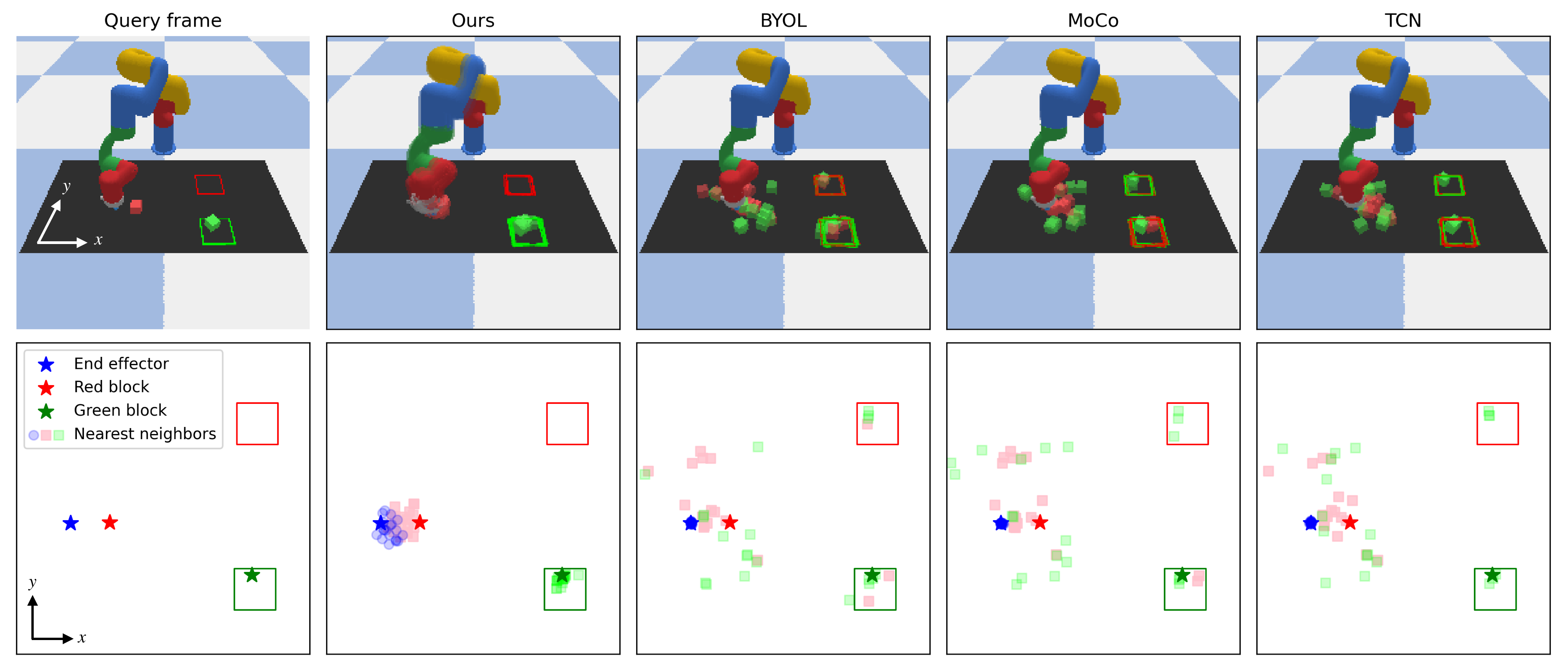}
  \caption{Embedding nearest neighbor matches for \method{}, BYOL, MoCo, and TCN on the Block Pushing environment. \textbf{(Top)} The nearest neighbor matches visualized in pixel space. \textbf{(Bottom)} Matches visualized in a top-down view. We see that the \method{} representation captures task-relevant features (end effector, block, and target locations in this case), whereas prior work fixates on the large robot arm.}
  \label{fig:blockpush_nn}
\end{figure}

\paragraph{Limitations of prior self-supervised techniques:} Prior self-supervised techniques can learn to fixate on visually salient features and ignore fine-grained features important for control. We illustrate this limitation using the Block Pushing environment from \citet{florence2022implicit}. In this task, the goal is to push a block into a target square. While the robot arm occupies much of the raw pixel space, the blocks are central to the task despite being smaller in the visual field. Figure \ref{fig:blockpush_nn} visualizes a random frame from the demonstration data and its $20$ nearest neighbors in the embedding space learned by several self-supervised techniques. 

We observe that prior self-supervised methods (details in \S \ref{exp:sim_results}) focus on the visually dominant robot, matching the whole robot arm extremely accurately. However, they fail to capture the block positions, which are important to the task despite being much less salient visually.

Can we learn a visual encoder that extracts task-specific features better? We know that the demonstrations are sequential: each observation is dependent on the previous observation, and an action (unobserved in this setting). Prior self-supervised methods ignore this sequential structure. Contrastive augmentations \cite{grill2020bootstrap, he2020momentum} and autoencoding objectives \cite{he2022masked, xiao2022masked, majumdar2024we} assume that the demonstration video is a bag of frames, discarding temporal information altogether. Temporal contrast \cite{young2022playful, sermanet2018time} uses temporal proximity but discards the sequential information in the observations: the contrastive objectives are usually symmetric in time, disregarding past/future order.

Instead of a contrastive or denoising objective, we propose a dynamics prediction objective that explicitly exploits the sequential structure of demonstration observations.

\begin{figure}[t]
  \centering
  \includegraphics[width=\textwidth]{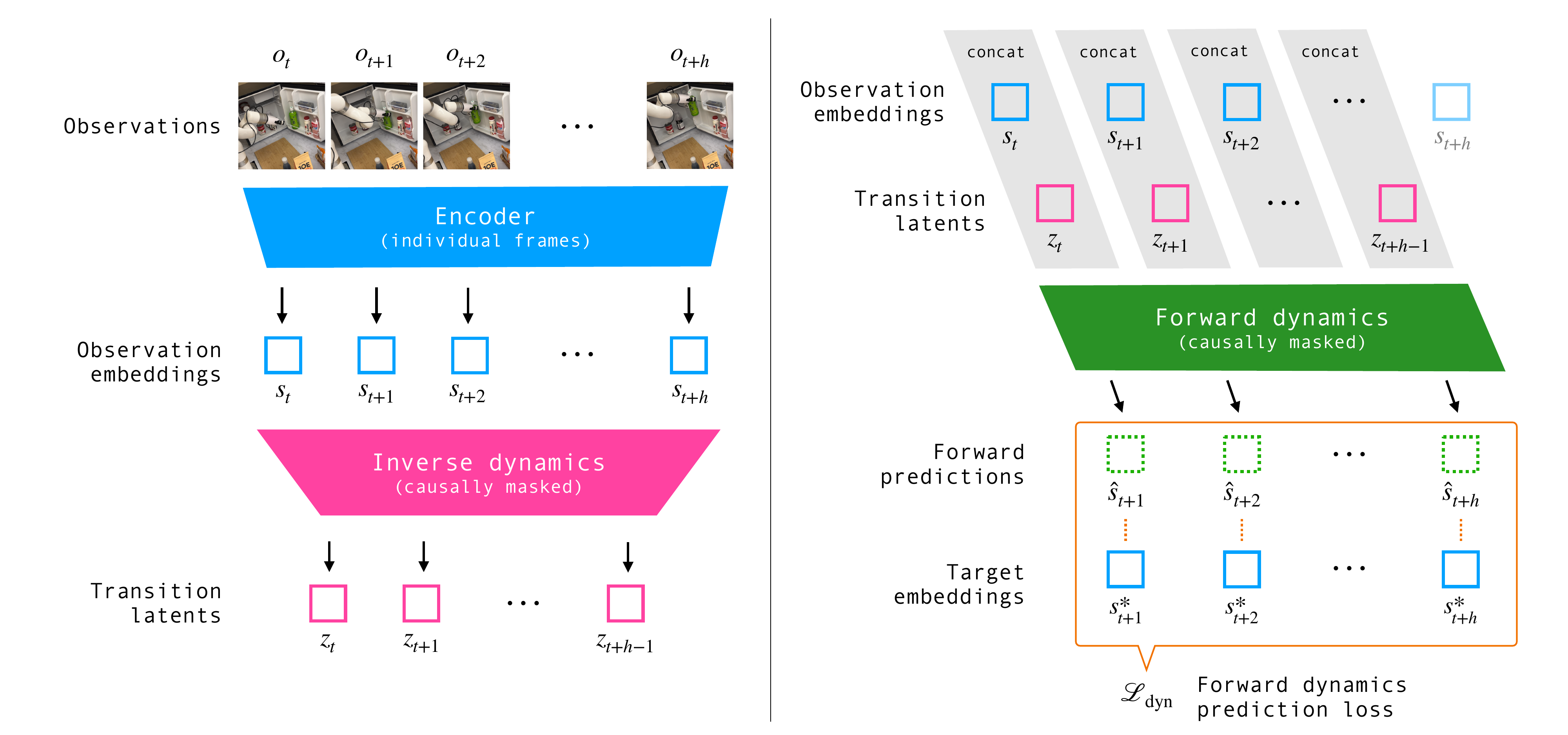}
  \caption{Architecture of \method{}. \method{} jointly learns an image encoder, an inverse dynamics model, and a forward dynamics model with a forward dynamics prediction loss.}
  \label{fig:arch}
\end{figure}

\paragraph{Overview of \method{}:} The key insight of our method is that we can learn a good visual representation for control by modeling the dynamics on demonstration observations, without requiring augmentations, contrastive sampling, or access to the ground truth actions. Given a sequence of raw visual observations $(o_1, \ldots, o_T)$, we jointly train the encoder $f_\theta: o_t \to s_t$, a latent inverse dynamics model $q(z_{t:t+h-1}|s_{t:t+h})$, and a forward dynamics model $p(\hat{s}_{t+1:t+h} | s_{t:t+h-1}, z_{t:t+h-1})$. We model the actions as unobserved latents, and train all models end-to-end with a consistency loss on the forward dynamics prediction. For our experiments, we use a ResNet18 \cite{he2016deep} encoder, and causally masked transformer encoders \cite{vaswani2017attention} for the inverse and forward dynamics models. The architecture is illustrated in Figure \ref{fig:arch}.

\subsection{Dynamics as a visual self-supervised learning objective}
\label{model:architecture}
First, we sample an observation sequence $o_{t:t+h}$ of length $h$ and compute its representation $s_{t:t+h} = f_\theta(o_{t:t+h})$. For convenience, we will write $s_{t:t+h}$ as $s_{:h}$, and $s_{t+1:t+h}$ as $s_{1:h}$ below. At any given step, the distribution of possible actions can be multimodal \cite{shafiullah2022behavior}. Therefore, the forward dynamics transition $p(s_{1:h} | s_{:h-1})$ can also have multiple modes. To address this, we first model the inverse dynamics $q(z_{:h-1} | s_{:h})$, where $z_t$ is the latent transition between frames. We assume $z_t$ to be well-determined and unimodal given consecutive frames $\{s_t, s_{t+1}\}$. We have $z \in \R^m, s \in \R^d, m \ll d$ such that the latent cannot trivially memorize the next frame embedding. Finally, we concatenate $(s_t, z_t)$ and predict the one-step forward dynamics $p(\hat{s}_{1:h} | s_{:h-1}, z_{:h-1})$.

We compute a dynamics loss $\mathcal{L}_{\mathrm{dyn}}(\hat{s}, s^*)$ on the one-step forward predictions $\hat{s}_{t+1:t+h}$, where $s^*_{t+1:t+h}$ are the target next-frame embeddings; and a covariance regularization loss $\mathcal{L}_{\mathrm{cov}}$ from \citet{bardes2021vicreg} on a minibatch of observation embeddings $S$:

\begin{equation}
\label{eq:objective}
\begin{aligned}
\mathcal{L}_{\mathrm{dyn}}(\hat{s}_t, s^*_t) &= 1 - \dfrac{\langle \hat{s}_t, s^*_t \rangle}{ \| \hat{s}_t \|_2 \cdot \| s^*_t \|_2} \\
\mathcal{L}_{\mathrm{cov}}(S) &= \frac{1}{d} \sum_{i \neq j} [\Cov(S)]^2_{i, j} \\
\mathcal{L} &= \mathcal{L}_\mathrm{dyn} + \lambda \mathcal{L}_\mathrm{cov}
\end{aligned}
\end{equation}

For environments with multiple views, we compute a loss over each view separately and take the mean. We choose $\lambda = 0.04$ following \citet{bardes2021vicreg} for the total loss $\mathcal{L}$. We find that covariance regularization slightly improves downstream task performance.

Naively, this objective admits a constant embedding solution. To prevent representation collapse, for $\mathcal{L}_\mathrm{dyn}(\hat{s}, s^*)$, we follow SimSiam~\cite{chen2021exploring} and set the target embedding $s^*_t := \mathrm{sg}(s_t)$, where $\mathrm{sg}$ is the stop gradient operator. Alternatively, our objective is also compatible with a target from a momentum encoder $f_{\bar\theta}$ \cite{he2020momentum, grill2020bootstrap}, $s^*_t := \bar{s}_t = f_{\bar\theta}(o_t)$, where $\bar\theta$ is an exponential moving average of $\theta$.

We train all three models end-to-end with the objective in Eq.~\ref{eq:objective}, and use the encoder for downstream control tasks.

\section{Experiments}
\label{sec:experiments}

We evaluate our dynamics-pretrained visual representation on a suite of simulated and real benchmarks. We compare \method{} representations with pretrained representations for vision and control, as well as other self-supervised learning methods. Our experiments are designed to answer the following questions:
\begin{enumerate*}[label=(\alph*)]
    \item Does \method{} improve downstream policy performance?
    \item Do representations trained with \method{} work on real robotic tasks?
    \item Is \method{} compatible with different policy classes?
    \item Can pretrained weights be fine-tuned in domain with \method{}?
    \item How important is each component in \method{}?
\end{enumerate*}

\subsection{Environments and datasets}
We evaluate \method{} on four simulated benchmarks and two real robot environments (depicted in Figure \ref{fig:envs}). We provide a brief description below with more details included in Appendix \ref{appdx: environment_details}.

\begin{enumerate}[label=(\alph*)]
    \item \textbf{Franka Kitchen} \cite{gupta2019relay}: The Franka Kitchen environment consists of seven simulated kitchen appliance manipulation tasks with a $9$-dimensional action space Franka arm and gripper. The dataset has $566$ demonstration trajectories, each completing three or four tasks. The observation space is RGB images of size $(224, 224)$ from a fixed viewpoint. We evaluate for 100 rollouts and report the mean number of completed tasks (maximum 4).

    \item \textbf{Block Pushing} \cite{florence2022implicit}: The simulated Block Pushing environment has two blocks, two target areas, and a robot pusher with $2$-dimensional action space (end-effector translation). Both the blocks and targets are colored red and green. The task is to push the blocks into either same- or opposite-colored targets. The dataset has $1\,000$ demonstration trajectories. The observation is RGB images of size $(224, 224)$ from two fixed viewpoints. We evaluate for 100 rollouts and report the mean number of blocks in targets (maximum 2).
    
    \item \textbf{Push-T} \cite{chi2023diffusion}: The environment consists of a pusher with $2$-dimensional action space, a T-shaped rigid block, and a target area in green. The task is to push the block to cover the target area. The dataset has $206$ demonstration trajectories. The observation space is a top-down view of the environment, rendered as RGB images of size $(224, 224)$. We evaluate for 100 rollouts and report the final coverage of the target area (maximum $1$).

    \item \textbf{LIBERO Goal} \cite{liu2024libero}: The environment consists of 10 manipulation tasks with a $7$-dimensional action space simulated Franka arm and gripper. The dataset has $500$ demonstration trajectories in total, $50$ per task goal. The observation space is RGB images of size $(224, 224)$ from a fixed external camera, and a wrist-mounted camera. We evaluate a goal-conditioned policy for $100$ rollouts in total, $10$ per task goal, and report the average success rate (maximum $1$).

    \item \textbf{Allegro Manipulation}: A real-world environment with an Allegro Hand attached to a Franka arm. We evaluate on three tasks: picking up a sponge ($6$ demonstrations), picking up a teabag ($7$ demonstrations), and opening a microwave ($6$ demonstrations). The observation space is RGB images of size $(224, 224)$ from a fixed external camera. The action space is $23$-dimensional, consisting of the Franka pose ($7$), and Allegro hand joint positions ($16$).
    
    \item \textbf{xArm Kitchen}: A real-world multi-task kitchen environment with an xArm robot arm and gripper. The environment consists of five manipulation tasks. The dataset includes $65$ demonstrations across five tasks. The observation space is RGB images of size $(128, 128)$ from three fixed external cameras, and an egocentric camera attached to the gripper. The action space is $7$-dimensional with the robot end effector pose and the gripper state.

\end{enumerate}

\subsection{Does \method{} improve downstream policy performance?}
\label{exp:sim_results}

We evaluate each representation by training an imitation policy head on the frozen embeddings, and reporting the downstream task performance on the simulated environments. 
We use Vector-Quantized Behavior Transformer (VQ-BeT) \cite{lee2024behavior} for the policy head. For xArm Kitchen, we use a goal-conditioned \textsc{Baku} \cite{haldar2024baku} with a VQ-BeT action head. MAE-style baselines (VC-1, MVP, MAE) use a ViT-B backbone. All other baselines and \method{} use a ResNet18 backbone.

For environments with multiple views, we concatenate the embeddings from all views for the downstream policy. Further training details are in Appendix~\ref{appdx:sec:hyperparams}. Table \ref{tab:sim_rep_results} provides comparisons of \method{} pretrained representations with other self-supervised learning methods, and pretrained weights for vision and robotic manipulation:

\begin{enumerate}[label=\textbullet]
    \item \textbf{Random, ImageNet, R3M}: ResNet18 with random, ImageNet-1K, and R3M \cite{nair2022r3m} weights.
    \item \textbf{VC-1}: Pretrained weights from \citet{majumdar2024we}.
    \item \textbf{MVP}: Pretrained weights from \citet{xiao2022masked}.
    \item \textbf{BYOL}: BYOL \cite{grill2020bootstrap} pretraining on demonstration data.
    \item \textbf{BYOL-T}: BYOL + temporal contrast \cite{young2022playful}. Adjacent frames $o_t, o_{t+1}$ are sampled as positive pairs, in addition to augmentations.
    \item \textbf{MoCo-v3}: MoCo \cite{he2020momentum} pretraining on demonstration data.
    \item \textbf{RPT}: RPT \cite{radosavovic2023robot} trained on observation tokens.
    \item \textbf{TCN}: Time-contrastive network \cite{sermanet2018time} pretraining on demonstrations. MV: multi-view objective; SV: single view objective.
    \item \textbf{MAE}: Masked autoencoder \cite{he2022masked} pretraining on demonstrations.
    \item \tb{\method{}}: \method{} pretraining on demonstrations.
\end{enumerate}

\begin{figure}[t]
  \centering
  \includegraphics[width=\textwidth]{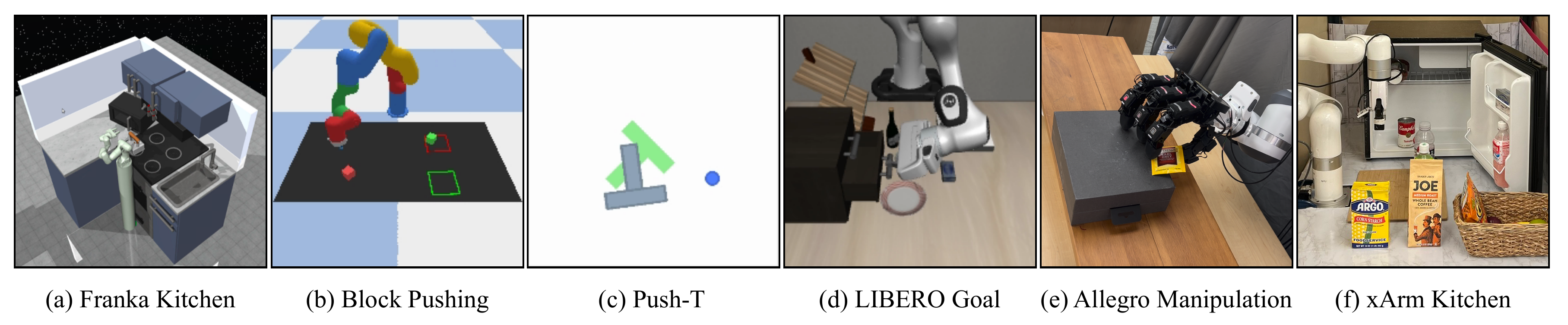}
  \caption{We evaluate \method{} on four simulated benchmarks - Franka Kitchen, Block Pushing, Push-T, and LIBERO Goal, and two real-world environments - Allegro Manipulation, and xArm Kitchen.}
  \label{fig:envs}
\end{figure}

\begin{table}[t]
\centering
\caption{Downstream policy performance on frozen visual representation on four simulated benchmarks - Franka Kitchen, Blocking Pushing, Push-T, and LIBERO Goal. We observe that \method{} matches or significantly outperforms prior work on all simulated tasks.}

\label{tab:sim_rep_results}
\begin{tabular}{@{}llcccc@{}}
\toprule
  & Method &
  \begin{tabular}[c]{@{}c@{}}Franka Kitchen \\ ( $\cdot / 4$ )\end{tabular} &
  \begin{tabular}[c]{@{}c@{}}Block Pushing  \\ ( $\cdot / 2$ )\end{tabular} &
  \begin{tabular}[c]{@{}c@{}}Push-T         \\ ( $\cdot / 1$ )\end{tabular} &
  \begin{tabular}[c]{@{}c@{}}LIBERO Goal    \\ ( $\cdot / 1$ )\end{tabular} \\
\midrule
                                                                                      & Random         & \tu{3.32} & 0.07       & 0.07      & 0.80      \\
\addlinespace[0.2ex]
\hdashline
\addlinespace[0.2ex]

\multirow{4}{*}{\begin{tabular}[c]{@{}l@{}}Pretrained\\ representations\end{tabular}} & ImageNet       & 3.01      & \tu{0.12}  & 0.41      & \tu{0.93} \\
                                                                                      & R3M            & 2.84      & 0.11       & \tu{0.49} & 0.89      \\
                                                                                      & VC-1           & 2.63      & 0.05       & 0.38      & 0.91      \\
                                                                                      & MVP            & 2.31      & 0.00       & 0.20      & 0.88      \\
                                                                                      \midrule
\multirow{6}{*}{\begin{tabular}[c]{@{}l@{}}Self-supervised\\ methods\end{tabular}}    & BYOL           & \tb{3.75} & 0.09       & 0.23      & 0.28      \\
                                                                                      & BYOL-T         & 3.33      & 0.16       & 0.34      & 0.28      \\
                                                                                      & MoCo-v3        & 3.28      & 0.03       & 0.57      & 0.70      \\
                                                                                      & RPT            & 3.54      & 0.52       & 0.56      & 0.17      \\
                                                                                      & TCN-MV         & ---       & 0.07       & ---       & 0.69      \\
                                                                                      & TCN-SV         & 2.41      & 0.07       & 0.07      & 0.76      \\
                                                                                      & MAE            & 2.70      & 0.00       & 0.07      & 0.59      \\
                                                                                      & \tb{\method{}} & 3.64      & \tb{0.65}  & \tb{0.66} & \tb{0.93} \\
\bottomrule
\end{tabular}
\end{table}

The best pretrained representation is \tu{underlined} and the best self-supervised representation is \tb{bolded}. We find that our method matches prior state-of-the-art visual representations on Franka Kitchen, and outperforms all other visual representations on Block Pushing, Push-T, and LIBERO Goal.

\subsection{Do representations trained with \method{} work on real robotic tasks?}
\label{exp:real_envs}

We evaluate the representations pre-trained with \method{} on two real-world robot environments: the Allegro Manipulation environment, and the multi-task xArm Kitchen environment. For the Allegro environment, we use a k-nearest neighbors policy \cite{pari2021surprising} and initialize with ImageNet-1K features for all pretraining methods, as the dataset is relatively small with around $1\,000$ frames per task. In the xArm Kitchen environment, we use the \textsc{Baku} \cite{haldar2024baku} architecture for goal-conditioned rollouts across five tasks. For our real-robot evaluations, we compare \method{} against the strongest performing baselines from our simulated experiments (see Table \ref{tab:sim_rep_results}). The results are reported in Table~\ref{tab:real_results}. We observe that \method{} outperforms the best baseline by 43\% on the single-task Allegro hand and by 20\% on the multi-task xArm Kitchen environment. Additionally, as shown in Table~\ref{tab:hand_pretrained}, \method{} exceeds the performance of pretrained representations by 50\% on the Allegro hand. These results demonstrate that \method{} is capable of learning effective robot representations in both single-task and multi-task settings.

\begin{table}[t]
\centering
\caption{We evaluate \method{} on eight tasks across two real-world environments: Allegro Manipulation, and xArm Kitchen. Results are presented as (successes/total). We observe that \method{} significantly outperforms prior representation learning methods on real tasks. }
\label{tab:real_results}

\begin{tabular}{l l c c c c}
\toprule
 & Task & BYOL & BYOL-T & MoCo-v3 & \tb{\method{}} \\
\midrule
\multirow{3}{*}{Allegro} & Sponge & 2/10 & 4/10 & 5/10 & \tb{7/10} \\
 & Tea & 1/10 & 0/10 & 2/10 & \tb{5/10} \\
 & Microwave & 2/10 & 3/10 & 1/10 & \tb{9/10} \\
\cmidrule(lr){1-6}
\multirow{5}{*}{xArm Kitchen} & Put yogurt & 4/5 & 4/5 & 2/5 & \tb{5/5} \\
 & Get yogurt & 0/5 & 4/5 & 4/5 & \tb{5/5} \\
 & Put ketchup & \tb{5/5} & 3/5 & \tb{5/5} & 4/5 \\
 & Get tea & 2/5 & 2/5 & 3/5 & \tb{5/5} \\
 & Get water & 0/5 & 0/5 & \tb{3/5} & \tb{3/5} \\
\bottomrule
\end{tabular}

\end{table}

\begin{wraptable}{R}{65mm}
\vspace{-12pt}
\caption{Pretrained baselines on Allegro}
\label{tab:hand_pretrained}
\begin{tabular}{l c c c}
\toprule
Method           & Sponge      & Tea       & Microwave \\
\midrule
ImageNet         & 4/10        & 1/10      & 0/10      \\
R3M              & 1/10        & 1/10      & 5/10      \\
\tb{\method{}}   & \tb{7/10}   & \tb{5/10} & \tb{9/10} \\
\bottomrule
\end{tabular}
\end{wraptable}

\subsection{Is \method{} compatible with different policy classes?}
On the Push-T environment~\cite{chi2023diffusion}, we compare all pretrained representations across four policy classes: VQ-BeT \cite{lee2024behavior}, Diffusion Policy \cite{chi2023diffusion}, MLP (with action chunking \cite{zhao2023learning}), and k-nearest neighbors with locally weighted regression \cite{pari2021surprising}. We present the results in Table \ref{tab:pusht_policies}. We find that \method{} representations improve downstream policy performance across policy classes compared to prior state-of-the-art representations. We also note that our representation works on the robot hand in \S \ref{exp:real_envs} with a nearest neighbor policy.

\begin{table}[t]
\centering
\caption{We evaluate the compatibility of \method{} with different policy classes for downstream policy learning on the Push-T simulated benchmark. We report the final target coverage achieved (maximum 1) and demonstrate that \method{} significantly outperforms prior representation learning methods across all policy classes.}
\label{tab:pusht_policies}
\begin{tabular}{@{}llcccc@{}}
\toprule
         &
  Method &
  VQ-BeT &
  Diffusion &
  MLP (chunking) &
  kNN \\
\midrule
                                                                                      & Random         & 0.07      & 0.04      & 0.07      & 0.01      \\
\addlinespace[0.2ex]
\hdashline
\addlinespace[0.2ex]
\multirow{4}{*}{\begin{tabular}[c]{@{}l@{}}Pretrained\\ representations\end{tabular}} & ImageNet       & 0.41      & \tb{0.73} & 0.24      & 0.09      \\
                                                                                      & R3M            & 0.49      & 0.63      & 0.27      & 0.08      \\
                                                                                      & VC-1           & 0.38      & 0.63      & 0.22      & 0.07      \\
                                                                                      & MVP            & 0.20      & 0.49      & 0.11      & 0.08      \\
                                                                                      \midrule
\multirow{6}{*}{\begin{tabular}[c]{@{}l@{}}Self-supervised\\ methods\end{tabular}}    & BYOL           & 0.23      & 0.40      & 0.11      & 0.04      \\
                                                                                      & BYOL-T         & 0.34      & 0.50      & 0.16      & 0.04      \\
                                                                                      & MoCo v3        & 0.57      & 0.67      & 0.30      & 0.07      \\
                                                                                      & RPT            & 0.56      & 0.62      & 0.30      & 0.07      \\
                                                                                      & TCN-SV         & 0.07      & 0.14      & 0.07      & 0.01      \\
                                                                                      & MAE            & 0.07      & 0.06      & 0.07      & 0.02      \\
                                                                                      & \tb{\method{}} & \tb{0.66} & \tb{0.73} & \tb{0.35} & \tb{0.12} \\
 \bottomrule
\end{tabular}
\end{table}

\subsection{Can pretrained weights be fine-tuned in domain with \method{}?}
We fine-tune an ImageNet-1K-pretrained ResNet18 with \method{} for each simulated environment, and evaluate with downstream policy performance on the frozen representation as described in \S \ref{exp:sim_results}. The results are shown in Table \ref{tab:imagenet_finetune}. We find that \method{} is compatible with ImageNet initialization, and can be used to fine-tune out-of-domain pretrained weights to further improve in-domain task performance. We also note that our method works in the low-data regime with ImageNet initialization on the real Allegro hand in Table \ref{tab:real_results}.

\begin{table}[t]
\centering
\caption{We evaluate the ability of \method{} to finetune an ImageNet-pretrained ResNet-18 encoder across 4 benchmarks. We demonstrate that using a pretrained encoder can further improve the performance of \method{}.}
\label{tab:imagenet_finetune}
\begin{tabular}{@{}lcccc@{}}
\toprule
  Representation &
  \multicolumn{1}{c}{\begin{tabular}[c]{@{}c@{}}Franka Kitchen \\ ( $\cdot / 4$ )\end{tabular}} &
  \multicolumn{1}{c}{\begin{tabular}[c]{@{}c@{}}Block Pushing  \\ ( $\cdot / 2$ )\end{tabular}} &
  \multicolumn{1}{c}{\begin{tabular}[c]{@{}c@{}}Push-T         \\ ( $\cdot / 1$ )\end{tabular}} &
  \multicolumn{1}{c}{\begin{tabular}[c]{@{}c@{}}LIBERO Goal    \\ ( $\cdot / 1$ )\end{tabular}} \\
\midrule
ImageNet                              & 3.01      & 0.12      & 0.41       & \tb{0.93}  \\
\tb{\method{}} (random init)          & 3.64      & 0.65      & \tb{0.66}  & \tb{0.93}  \\
\tb{\method{}} (ImageNet fine-tuned)  & \tb{3.82} & \tb{0.67} & 0.50       & 0.90       \\ 
\bottomrule
\end{tabular}
\end{table}

\subsection{How important is each component in \method{}?}
\label{exp:ablations}

In Table~\ref{tab:ablations}, we ablate each component in \method{} and measure its impact on downstream policy performance on our simulated benchmarks.

\begin{table}[t]
\centering
\caption{Ablation analysis of downstream performance relative to the full architecture (100\%)}
\label{tab:ablations}

\begin{tabular}{@{}lrrrr@{}}
\toprule
Ablations     & Kitchen & Block & Push-T & LIBERO \\
\midrule
No forward    &  34\%   &  8\%  & 44\%   & 33\%   \\
No inverse    &  72\%   & 35\%  & 97\%   & 41\%   \\
No bottleneck &  92\%   & 22\%  &  9\%   & 75\%   \\
No cov. reg.  &  94\%   & 62\%  & 85\%   & 59\%   \\
No stop grad. &   1\%   &  5\%  &  9\%   & 0\%    \\
Short context & 100\%   & 75\%  & 88\%   & 89\%   \\ 
\bottomrule
\end{tabular}
\end{table}

\textbf{Forward dynamics prediction}: We replace the one-step forward prediction target $s^*_{1:h}$ with the same-step target $s^*_{:h-1}$. To prevent the model from trivially predicting $s^*_t$ given $s_t$, we replace the forward dynamics input $(s_{:h-1}, z_{:h-1})$ with only $z_{:h-1}$. The ablated objective is essentially a variant of autoencoding $s_t$. We observe that removing forward dynamics prediction degrades performance across environments.

\textbf{Inverse dynamics to a transition latent}: As described in \S \ref{model:architecture}, the forward dynamics loss assumes that the transition is unimodal and requires an inferred transition latent. We observed that removing the latent from the forward dynamics input results in a significant performance drop.

\textbf{Bottleneck on the transition latent dimension}: For the transition latent $z$ and the observation embedding $s$, we find that having $\dim z \ll \dim s$ stabilizes training. Here we set $\dim z := \dim s$, and find that our model can still learn a reasonable representation in some environments, but training can destabilize, leading to a high variance in downstream performance.

\textbf{Covariance regularization}: We find that covariance regularization from \citet{bardes2021vicreg} improves performance across environments. Training still converges without it, but the downstream performance is slightly worse.

\textbf{Stop gradient on target embeddings}: We observe that removing techniques like momentum encoder \cite{he2020momentum, grill2020bootstrap} and stop gradient \cite{chen2021exploring} leads to representation collapse~\cite{chen2020simple, grill2020bootstrap, bardes2021vicreg}.

\textbf{Observation context}: The dynamics objective requires at least $2$ frames of observation context. For Franka Kitchen, we find that a context of $2$ frames works best. For the other environments, a longer observation context ($5$ frames) improves downstream policy performance. Details of hyperparameters used for \method{} visual pretraining can be found in Appendix \ref{appdx: visual-encoder-training}.

\subsection{Variants with access to ground truth actions}
In Table \ref{tab:variants-with-gt-actions}, we compare with two variants of \method{} where we assume access to ground truth action labels during visual encoder training.

\begin{table}[t]
\centering
\caption{Variants with ground truth actions, downstream performance relative to the base model (100\%)}
\label{tab:variants-with-gt-actions}

\begin{tabular}{@{}lrrrr@{}}
\toprule
Variants                          & Kitchen & Block & Push-T & LIBERO \\
\midrule
Inverse dynamics only             &  100\%  & 54\%  & 70\%   & 11\%   \\
\method{} + action labels         &  97\%   & 29\%  & 94\%   & 86\%   \\
\bottomrule
\end{tabular}
\end{table}

\tb{Only inverse dynamics to ground truth actions}: as proposed in \citet{brandfonbrener2024inverse}, we train the visual encoder by learning an inverse dynamics model to ground truth actions, with covariance regularization, and without forward dynamics.

\tb{Full model + inverse dynamics to ground truth actions}: we train the full \method{} model plus an MLP head to predict the ground truth actions given the transition latents inferred by the inverse dynamics model.

We observe that in both cases, having access to ground truth actions during visual pretraining does not seem to improve downstream policy performance. We hypothesize that this is because the downstream policy already has access to the same actions for imitation learning.
\section{Related works}
\label{sec:related_works}

This work builds on a large body of research on self-supervised visual representations, learning from human demonstrations, neuroscientific basis for learning dynamics for control, predictive models for decision making, learning from videos for control, and visual pretraining for control.

\paragraph{Self-supervised visual representations:} Self-supervised visual representations have been widely studied since the inception of deep learning. There are several common approaches to self-supervised visual representation learning. One approach is to recover the ground truth from noise-degraded samples using techniques like denoising autoencoders \cite{xiang2023denoising, sterzentsenko2019self} and masked modeling \cite{devlin2018bert, bao2021beit, he2022masked}. Another approach is contrastive learning, which leverages data augmentation priors \cite{chen2020simple, grill2020bootstrap, he2020momentum, bardes2021vicreg, chen2021exploring} or temporal proximity \cite{sermanet2018time, oord2018representation} to produce contrastive sample pairs. A third self-supervised method is generative modeling \cite{chen2020generative, van2016pixel, trinh2019selfie}, which learns to sequentially generate the ground truth data. More recently, self-supervision in the latent space rather than the raw pixel space has proven effective, as seen in methods that predict representations in latent space \cite{assran2023self, bardes2023v}. 

\paragraph{Learning from demonstrations:} Learning from human demonstrations is a well-established idea in robotics \cite{delson1996robot, kaiser1996building, liu1993teaching, asada1990skill}. With the advances in deep learning, recent works such as \cite{chi2023diffusion, zhao2023learning, shafiullah2022behavior, cui2022play, lee2024behavior, reuss2023goal} show that imitation learning from human demonstrations has become a viable approach for training robotic policies in simulated and real-world settings.

\paragraph{Neural basis for learning dynamics:} It is widely believed that animals possess internal dynamics models that facilitate motor control. These models learn representations that are predictive of sensory inputs for decision making and motor control \cite{sutton1981toward, von1867handbuch, bastos2012canonical, barrett2015interoceptive}. Early works such as \cite{wolpert1995internal, wolpert1998internal, shidara1993inverse, kitazawa1998cerebellar} propose that there exists an internal model of the motor apparatus in the cerebellum for motor control and planning. \cite{miall1996forward, jordan1992forward} propose that the central nervous system uses forward models that predict motor command outcomes and model the environment. Learning forward and inverse dynamics models also helps with generalization to diverse task conditions \cite{flanagan1997role, haruno1998multiple}.

\paragraph{Predictive models for decision making:} Predictive model learning for decision making is well-established in machine learning. Learning generative models that can predict sequential inputs has achieved success across many domains, such as natural language processing \cite{radford2019language}, reinforcement learning \cite{seo2022reinforcement, schwarzer2020data, schwarzer2021pretraining}, and representation learning \cite{oord2018representation,schmeckpeper2020learning}. Incorporating the prediction of future states as an intrinsic reward has also been shown to improve reinforcement learning performance \cite{pathak2017curiosity, shelhamer2016loss, guo2022byol}. Moreover, recent work demonstrates that world models trained to predict environment dynamics can enable planning in complex tasks and environments \cite{schrittwieser2020mastering, hafner2019learning, hafner2019dream, hafner2020mastering, bruce2024genie}.

\paragraph{Learning from video for control:} Videos provide rich spatiotemporal information that can be leveraged for self-supervised representation learning~ \cite{goroshin2015unsupervised, bardes2024revisiting, feichtenhofer2021large, wang2019learning, dwibedi2019temporal, pirk2019online}. These methods have been extended to decision-making through effective downstream policy learning~\cite{ma2022vip, xiao2022masked, nair2022r3m, parisi2022unsurprising, majumdar2024we, chen2021learning}. Further, recent work also enables learning robotic policies directly from in-domain human demonstration videos by incorporating some additional priors~\cite{bahl2022human, sharma2018multiple, chen2021unsupervised, qin2022dexmv, sivakumar2022robotic}, as well as learning behavioral priors from actionless demonstration data \cite{edwards2019imitating, schmidt2023learning, ye2022become}.

\paragraph{Visual representation for control:} Visual representation learning for control has been an active area of research. Prior work has shown that data augmentation improves the robustness of learned representations and policy performance in reinforcement learning domains~\cite{kostrikov2020image, laskin2020reinforcement}. Additionally, pretraining visual representations on large out-of-domain datasets before fine-tuning for control tasks has been shown to outperform training policies from scratch~\cite{parisi2022unsurprising, dasari2023unbiased, nair2022r3m, majumdar2024we, radosavovic2023real, xiao2022masked, padalkar2023open}. More recent work has shown that in-domain self-supervised pretraining improves policy performance \cite{shafiullah2023bringing, zhou2023train, guzey2023dexterity, zheng2024premier} and enables non-parametric downstream policies \cite{pari2021surprising}.

\section{Discussion and Limitations}
\label{sec:discussion}

In this work, we have presented \method{}, a self-supervised algorithm for robot representation learning that leverages the sequential nature of demonstration data. \method{} incorporates predictive dynamics modeling to learn visual features that capture the sequential structure of demonstration observations. During pretraining, \method{} jointly optimizes the visual encoder with dynamics models to extract task-specific features. These learned representations can then be used for downstream control tasks, leading to more efficient policy learning compared to prior approaches. We believe that training \method{} on larger unlabeled datasets could potentially improve generalization. Additionally, while promising for control tasks, more research is needed to evaluate \method{}'s effectiveness on robotic manipulation outside of lab settings.
\section*{Acknowledgements}
We would like to thank
Ademi Adeniji,
Alex Wang,
Gaoyue Zhou,
Haritheja Etukuru,
Irmak Güzey,
Mahi Shafiullah,
Nikhil Bhattasali,
Raunaq Bhirangi,
Seungjae (Jay) Lee,
and Ulyana Piterbarg
for their valuable feedback and discussions. This work was supported by grants from Honda, Google, NSF award 2339096 and ONR awards N00014-21-1-2758 and N00014-22-1-2773. LP is supported by the Packard Fellowship.

\bibliography{references}
\bibliographystyle{unsrtnat}

%%%%%%%%%%%%%%%%%%%%%%%%%%%%%%%%%%%%%%%%%%%%%%%%%%%%%%%%%%%%

\newpage
\appendix

\section{Environment and dataset details}
\label{appdx: environment_details}

\subsection{Franka Kitchen}
The Franka Kitchen environment introdued by \citet{gupta2019relay} consists of a Franka arm with a $9$-dimensional action space. This environment includes seven tasks and a dataset of $566$ human-collected demonstrations. While the original environment is state-based, we created an image-based variant by rendering the states to $224 \times 224$ RGB images.

\subsection{Block Pushing}
In the Block Pushing environment introduced by \citet{florence2022implicit}, the objective is for the robot to push two colored blocks (red and green) into two target squares (also red and green). The training dataset consists of $1\,000$ trajectories, evenly distributed among the four possible combinations of block target and push order. These trajectories were collected by a scripted expert controller.

\subsection{Push-T}
In the Push-T environment introduced by \citet{chi2023diffusion}, the goal is to push a T-shaped block to a designated target position on a table. The dataset for this environment contains $206$ demonstrations collected by human operators. The action space in this environment is a two-dimensional end-effector position control. Similar to the Franka Kitchen environment, we have created an image-based variant by rendering demonstrations to $224 \times 224$ RGB images.

\subsection{LIBERO Goal}
In the LIBERO Goal environment introduced by \citet{liu2024libero}, there are 10 manipulation tasks, each with $50$ teleoperated demonstrations for goal-conditioned policy benchmarking. The environment has a $7$-dimensional action space and an observation space of $224 \times 224$ RGB images from two cameras (fixed external view, and wrist-mounted egocentric view).

\subsection{Allegro Manipulation}
The environment consists of an Allegro hand attached to a Franka arm, and a fixed camera for image observations. The observation space is $224 \times 224$ RGB images. The action space is $23$-dimensional, consisting of Cartesian position and orientation of the Franka robot arm (7 DoF), and 16 joint positions of the Allegro Robot Hand. The demonstrations are collected at 50Hz for Franka, and 60Hz for the Allegro hand. The learned policies are rolled out at 4Hz.

We evaluate on three contact-rich dexterous manipulation tasks that require precise multi-finger control and arm movement, described in detail below. 

\tb{Sponge picking}: This task requires the hand to reach to the position of the sponge, grasp the sponge, and lift the sponge from the table. We collect $6$ demonstrations via OpenTeach \cite{iyer2024open} for the task, starting from different positions, with $543$ frames in total. The task is considered successful if the robot hand can grasp the sponge from the table within 120 seconds.

\tb{Teabag picking}: This task is similar to the previous task, but more difficult with a smaller task object. We collect $7$ demonstrations via OpenTeach with $1\,034$ frames in total. In this task, the robot needs reach the teabag, grasp the teabag with two fingers, then pick it up. The task is considered successful if the robot hand can grasp the teabag from the table within 240 seconds.

\tb{Microwave opening}: This task requires the hand to reach the microwave door handle, grasp the handle, and pull down the door. We collect $6$ demonstrations via OpenTeach with $735$ frames in total. The task is considered successful if the robot hand can open the door within 240 seconds.

\subsection{xArm Kitchen}
\begin{figure}[!th]
    \centering
    \includegraphics[width=\textwidth]{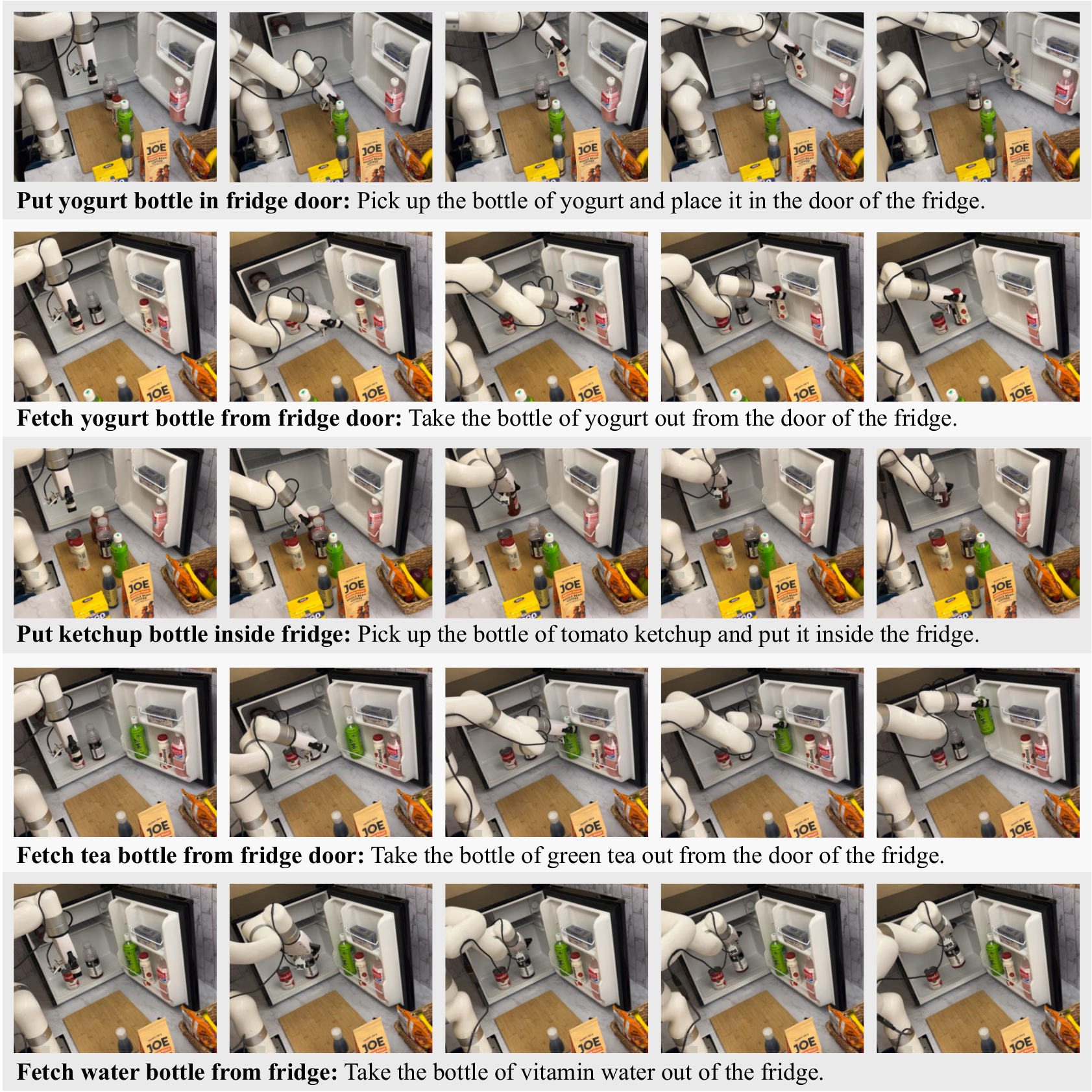}
    \caption{xArm Kitchen environment tasks}
    \label{fig:real_tasks}
\end{figure}

This is a real-world multi-task kitchen environment comprising a Ufactory xArm 7 robot with an xArm Gripper. The policies are trained on RGB images of size $128\times 128$ obtained from four different camera views, including an egocentric camera attached to the robot gripper. The action space comprises the robot end effector pose and the gripper state.  We collect a total of 65 demonstrations across 5 tasks, depicted in Figure \ref{fig:real_tasks}. The demonstrations were collected using OpenTeach \cite{iyer2024open} at 30Hz. The learned policies are deployed at 10Hz. Figure \ref{fig:real_tasks} shows real-world task rollouts for the multitask policy learned for all 5 tasks.

\newpage
\section{Hyperparameters and implementation details}
\label{appdx:sec:hyperparams}
\subsection{Visual encoder training}
\label{appdx: visual-encoder-training}
We present the \method{} hyperparameters below.
\begin{table}[h]
\centering
\caption{Environment-dependent hyperparameters for DynaMo pretraining, random init}
\label{tab:dynamo_hyperparams}
\begin{tabular}{@{}lcccc@{}}
\toprule
                  & Obs. context & EMA $\beta$   & Forward dynamics dropout & Transition latent dim \\
\midrule
Franka Kitchen    & 2            & SimSiam       & 0                        & 64                    \\
Block Pushing     & 5            & 0.99          & 0.3                      & 16                    \\
Push-T            & 5            & SimSiam       & 0                        & 8                     \\
LIBERO Goal       & 5            & SimSiam       & 0                        & 32                    \\
xArm Kitchen      & 5            & 0.99          & 0                        & 64                    \\
\bottomrule
\end{tabular}
\end{table}

\begin{table}[h]
\centering
\caption{Shared hyperparameters for DynaMo pretraining, random init}
\label{tab:dynamo_shared_hyperparameters}
\begin{tabular}{@{}lc@{}}
\toprule
Name                        & Value        \\
\midrule
Optimizer                   & AdamW        \\
Learning rate               & $10^{-4}$    \\
Weight decay                & 0.0          \\
Betas                       & (0.9, 0.999) \\
Gradient clip norm          & 0.1          \\
Covariance reg. coefficient & 0.04         \\
Epochs                      & 40           \\
Batch size                  & 64           \\
\bottomrule
\end{tabular}
\end{table}

\begin{table}[h]
\centering
\caption{Environment-dependent hyperparameters for DynaMo fine-tuning from ImageNet weights}
\label{tab:dynamo_finetune_hyperparams}
\begin{tabular}{@{}lccc@{}}
\toprule
                  & Obs. context & EMA $\beta$               & Transition latent dim \\
\midrule
Franka Kitchen    & 2            & SimSiam                   & 64                    \\
Block Pushing     & 5            & 0.99                      & 16                    \\
Push-T            & 5            & SimSiam                   & 8                     \\
LIBERO Goal       & 5            & 0.99                      & 32                    \\
Allegro           & 5            & SimSiam                   & 32                    \\
\bottomrule
\end{tabular}
\end{table}

\begin{table}[h!]
\centering
\caption{Shared hyperparameters for DynaMo fine-tuning}
\label{tab:dynamo_shared_finetuning_hyperparameters}
\begin{tabular}{@{}lc@{}}
\toprule
Name                        & Value        \\
\midrule
Optimizer                   & AdamW        \\
Learning rate               & $10^{-5}$    \\
Forward dynamics dropout    & 0.0          \\
Weight decay                & 0.0          \\
Betas                       & (0.9, 0.999) \\
Gradient clip norm          & 0.1          \\
Covariance reg. coefficient & 0.04         \\
Epochs                      & 40           \\
Batch size                  & 64           \\
\bottomrule
\end{tabular}
\end{table}

For Block Pushing and xArm kitchen, we use an EMA encoder with the beta schedule from the MoCo-v3 official repo. For \method{} training, we use a constant learning rate schedule for LIBERO Goal, and a cosine learning rate decay schedule with 5 warmup epochs on all other environments. For \method{} fine-tuning, we use a cosine learning rate decay schedule with 5 warmup epochs on all environments.

We use the following official implementation repos:
\begin{enumerate}[label=\textbullet]
  \item MoCo-v3: \url{https://github.com/facebookresearch/moco-v3}
  \item BYOL: \url{https://github.com/lucidrains/byol-pytorch}
  \item MAE: \url{https://github.com/facebookresearch/mae}
  \item R3M: \url{https://github.com/facebookresearch/r3m/}
  \item MVP: \url{https://github.com/ir413/mvp}
  \item VC-1: \url{https://github.com/facebookresearch/eai-vc}
\end{enumerate}

We base our transformer encoder implementation on nanoGPT \cite{karpathy2023nanogpt} at \url{https://github.com/karpathy/nanoGPT}.

For the Allegro Manipulation environment, we fine-tune MoCo and BYOL from ImageNet-1K weights for $1\,000$ epochs. For all other environments, we train MoCo and BYOL for $200$ epochs, MAE for 400 epochs, all from random initialization. The hyperparameters used for training these models are detailed in Table \ref{tab:moco_byol}.

Compute used for training \method{}:
\begin{enumerate}[label=\textbullet]
    \item Franka Kitchen: 3 hours on 1x NVIDIA A100.
    \item Block Pushing: 7 hours on 1x NVIDIA A100.
    \item Push-T: 1 hour on 1x NVIDIA A100.
    \item LIBERO Goal: 2 hours on 1x NVIDIA H100.
    \item Allegro Manipulation: 3 minutes on 1x NVIDIA RTX A6000 for the sponge task, 4 minutes for the teabag task, and 3 minutes for the microwave task.
    \item xArm kitchen: 4 hours on 1x NVIDIA RTX A6000.
\end{enumerate}

\begin{table}[h]
\centering
\caption{SSL Hyperparameters}
\label{tab:moco_byol}
\begin{minipage}{.33\linewidth}
\centering
\subcaption{MoCo Hyperparameters}
\label{tab:moco}
\begin{tabular}{@{}lc@{}}
\toprule
Name               & Value      \\
\midrule
Optimizer          & LARS  \\
Batch size         & 1024       \\
Learning rate      & 0.6        \\
Momentum           & 0.9        \\
Weight decay       & $10^{-6}$  \\
\bottomrule
\end{tabular}
\end{minipage}
\begin{minipage}{.33\linewidth}
\centering
\subcaption{BYOL Hyperparameters}
\label{tab:byol}
\begin{tabular}{@{}lc@{}}
\toprule
Name                & Value                 \\
\midrule
Optimizer           & LARS            \\
Batch size          & 512                   \\
Learning rate       & 0.2                   \\
Momentum            & 0.9                   \\
Weight decay        & $1.5 \times 10^{-6}$  \\
\bottomrule
\end{tabular}
\end{minipage}
\begin{minipage}{.33\linewidth}
\centering
\subcaption{MAE Hyperparameters}
\label{tab:simclr}
\begin{tabular}{@{}lc@{}}
\toprule
Name                & Value                 \\
\midrule
Optimizer           & AdamW                \\
Batch size          & 64                   \\
Learning rate       & $2.5 \times 10^{-5}$ \\
Weight decay        & 0.05   \\
\bottomrule
\end{tabular}
\end{minipage}
\end{table}

\subsection{Downstream policy training}
\label{appdx: policy-training}

Table \ref{tab:vq_bet_hyperparams}, \ref{tab:diffusion_hyperparams} and \ref{tab:mlp_hyperparams} detail the downstream policy hyperparameters for VQ-BeT, Diffusion Policy and MLP training for the simulated environments. 

For VQ-BeT, we use the implementation from the original paper \cite{lee2024behavior} with the recommended hyperparameters. For Diffusion Policy, we use the implementation at \url{https://github.com/real-stanford/diffusion_policy} with a transformer-based noise prediction network with the recommended hyperparameters. We use AdamW as optimizer for the three policy heads.

Compute used for downstream policy training:
\begin{enumerate}[label=\textbullet]
\item Franka Kitchen VQ-BeT: 8.5 hours on 1x NVIDIA A4000.
\item Block Pushing VQ-BeT: 4 hours on 1x NVIDIA A100.
\item Push-T VQ-BeT: 7 hours on 1x NVIDIA A100.
\item Push-T Diffusion Policy: 8 hours on 1x NVIDIA A100.
\item Push-T MLP: 2 hours on 1x NVIDIA A100.
\item LIBERO Goal VQ-BeT: 5 hours on 1x NVIDIA A4000.
\item xArm Kitchen VQ-BeT: 6 hours on 1x NVIDIA A4000.
\end{enumerate}

\begin{table}[h]
\centering
\caption{Hyperparameters for VQ-BeT training}
\label{tab:vq_bet_hyperparams}
\begin{tabular}{@{}lccccc@{}}
\toprule
Parameter              & Franka Kitchen & Block Pushing & Push-T  & LIBERO Goal\\
\midrule
Batch size             & 2048             & 64            & 512   & 64\\
Epochs                 & 1000            & 300           & 5000   & 50\\
Window size            & 10              & 3             & 5      & 10\\
Prediction window size & 1              & 1             & 5       & 1\\
Learning rate          & $5.5 \times 10^{-5}$      & $10^{-4}$     & $5.5 \times 10^{-5}$ & $5.5 \times 10^{-5}$\\
Weight decay           & $2\times10^{-4}$              & 0             & $2\times10^{-4}$ & $2\times10^{-4}$\\
\bottomrule
\end{tabular}
\end{table}

\begin{table}[h]
\centering
\caption{Hyperparameters for Diffusion Policy Training}
\label{tab:diffusion_hyperparams}
\begin{tabular}{@{}lcc@{}}
\toprule
Parameter           & Push-T \\
\midrule
Batch size          & 256    \\
Epochs              & 2000   \\
Learning rate       & $10^{-4}$ \\
Weight decay        & 0      \\
Observation horizon & 2      \\
Prediction horizon  & 10     \\
Action horizon      & 8      \\
\bottomrule
\end{tabular}
\end{table}
\begin{table}[h]
\centering
\caption{Hyperparameters for MLP Training}
\label{tab:mlp_hyperparams}
\begin{tabular}{@{}lcc@{}}
\toprule
Parameter            & Push-T \\
\midrule
Batch size           & 256    \\
Epochs               & 2000   \\
Learning rate        & $10^{-4}$ \\
Weight decay         & 0      \\
Hidden dim           & 256    \\
Hidden depth         & 8      \\
Observation context  & 5      \\
Prediction context   & 5      \\
\bottomrule
\end{tabular}
\end{table}

\newpage
\section{Real robot environment rollouts}

\begin{figure}[!h]
  \centering
  \includegraphics[width=\textwidth]{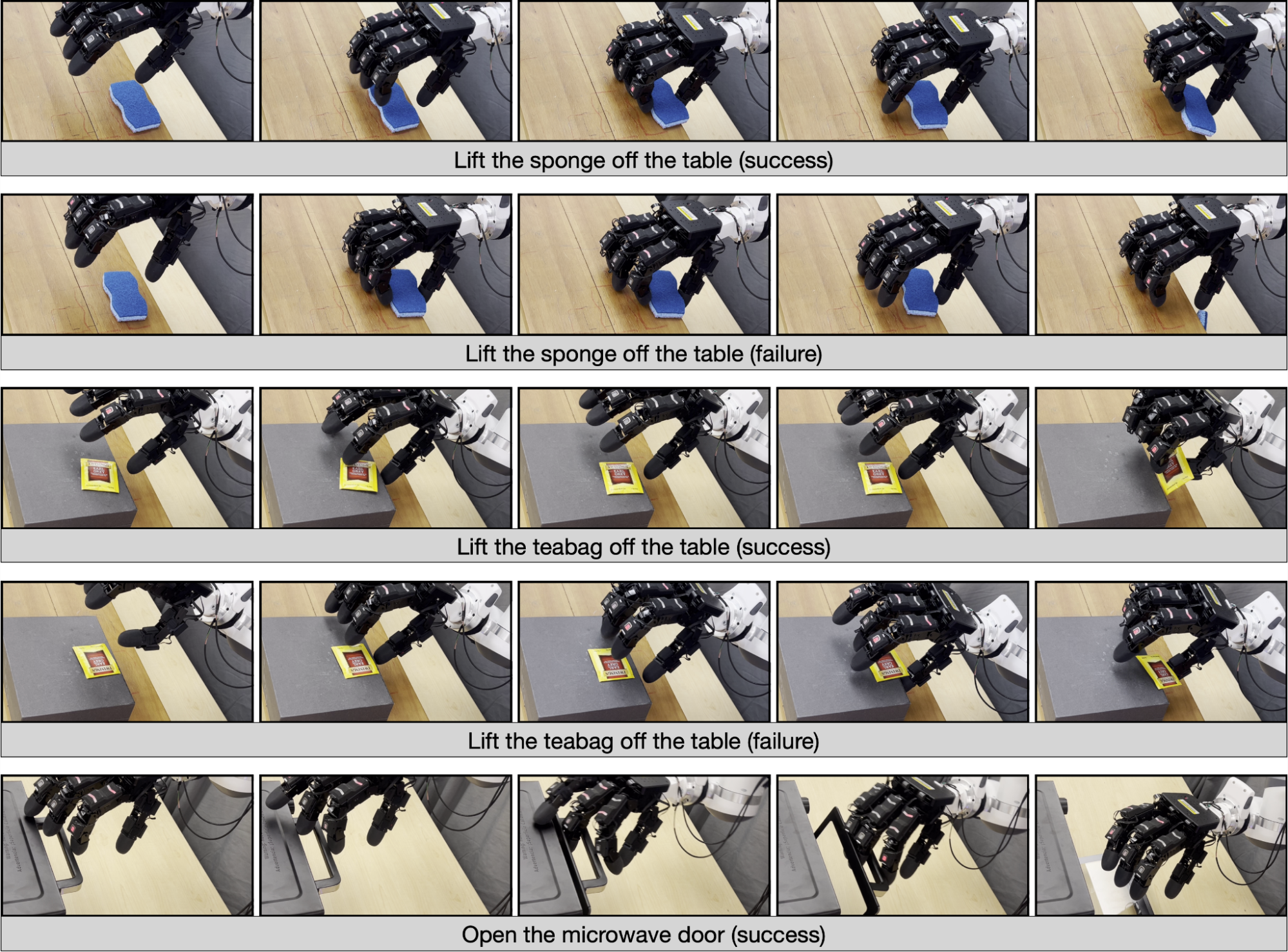}
  \caption{Rollouts on Allegro Manipulation with our \method{}-pretrained encoder.}
  \label{fig:allegro_rollouts}
\end{figure}

\newpage
\begin{figure}[!h]
  \centering
  \includegraphics[width=\textwidth]{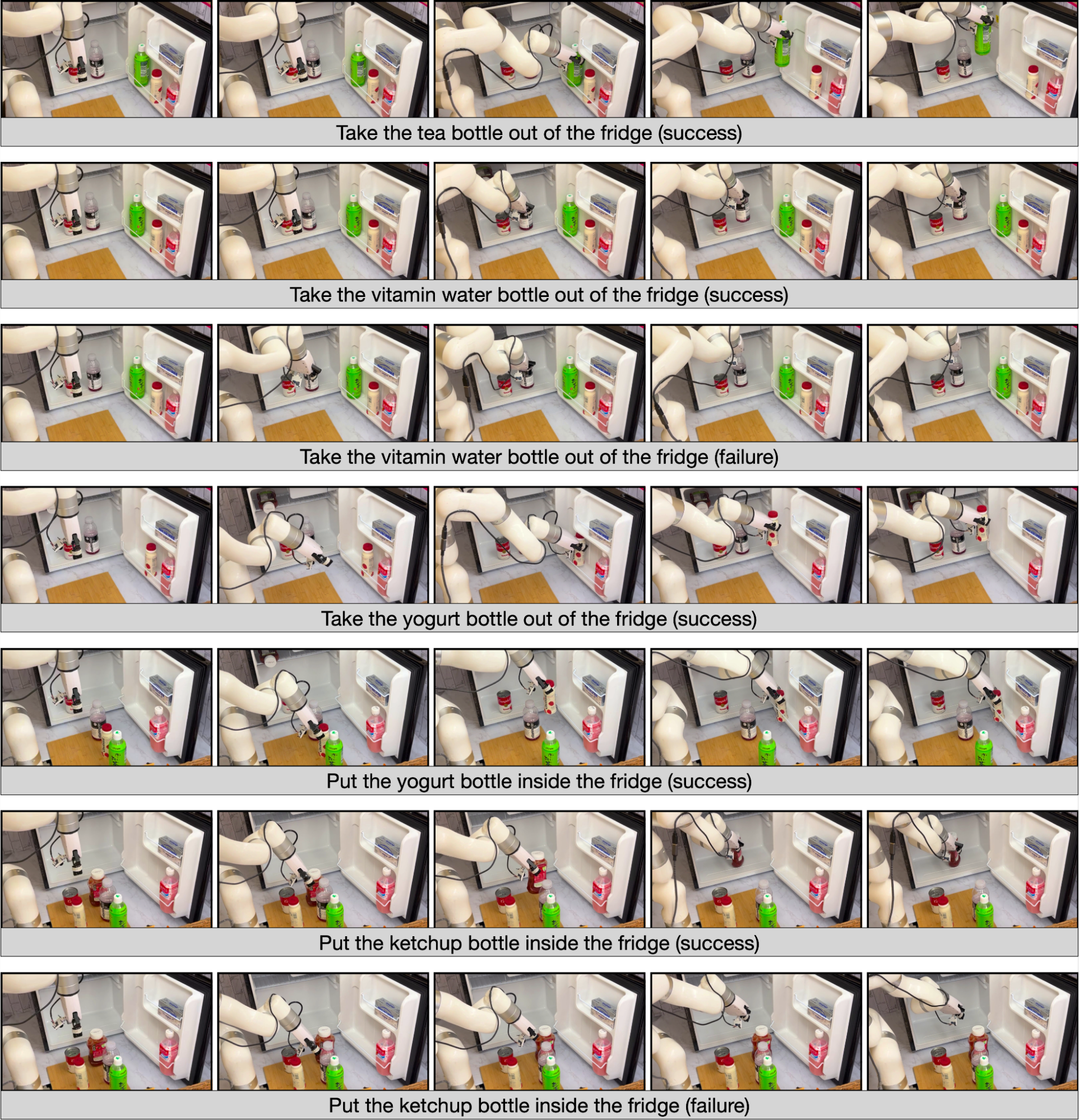}
  \caption{Rollouts on xArm Kitchen with our \method{}-pretrained encoder.}
  \label{fig:xarm_rollouts}
\end{figure}

%%%%%%%%%%%%%%%%%%%%%%%%%%%%%%%%%%%%%%%%%%%%%%%%%%%%%%%%%%%%

\end{document}